\documentclass[12pt]{article}

\usepackage[english]{babel}

\usepackage{times}
\usepackage{geometry}
\usepackage[utf8]{inputenc}
\usepackage{enumitem,amssymb}
\usepackage{ragged2e}
\usepackage{multicol}
\usepackage{multirow}
\usepackage{xcolor}
\usepackage{enumitem}
\usepackage{graphicx}
\usepackage{booktabs}
\usepackage{amsmath,amssymb,amsfonts,amsthm}

\usepackage{url}

\usepackage{amsmath,amsfonts,bm}

\def\eqref#1{equation~\ref{#1}}

\def\1{\bm{1}}

\DeclareMathAlphabet{\mathsfit}{\encodingdefault}{\sfdefault}{m}{sl}
\SetMathAlphabet{\mathsfit}{bold}{\encodingdefault}{\sfdefault}{bx}{n}

\usepackage{url}

\usepackage{graphicx}
\usepackage{booktabs} \usepackage{hyperref}

\usepackage{caption}

\usepackage{amsmath,amssymb,amsfonts,amsthm}

\usepackage{xcolor}
\usepackage{mathtools}
\usepackage{dsfont}
\usepackage{hyperref}
\usepackage{bm}
\usepackage{nicefrac}
\usepackage{wrapfig}
\usepackage{lipsum}

\usepackage{algorithm,algorithmic}
\usepackage[linesnumbered, ruled,vlined,algo2e]{algorithm2e}

\newcounter{assumption}\renewcommand{\theassumption}{\arabic{assumption}}

\def\rT{{\rm T}}

\def\normal{\mathcal{N}}

\newcommand*\lrbb[1]{\left\{#1\right\}}
\newcommand*\lrp[1]{\left(#1\right)}

\newcommand*\ind[1]{{\mathds{1}\lrbb{#1}}}

\def\rT{{\rm T}}

\newcommand{\NORMAL}{\ensuremath{\mathcal{N}}}

\usepackage{multirow}
\usepackage{caption}
\usepackage{subcaption}

\newcommand{\ts}{\textsuperscript}

\newlist{thematic}{itemize}{8}
\setlist[thematic]{label=$\square$}

  \newcommand{\V}[1]{{\mathbf{#1}}}

\begin{document}
\raggedright
\huge
DeepGLEAM: A hybrid mechanistic and deep learning model for COVID-19 forecasting

\normalsize

\textbf{Authors:} \\
Dongxia (Allen) Wu $^\dagger$, Liyao Gao, Xinyue Xiong $^\star$,\\
Matteo Chinazzi$^\star$, Alessandro Vespignani$^\star$, Yi-An Ma$^\dagger$, Rose Yu$^\dagger$\\

\textbf{Institution:} \\
$^\dagger$ University of California, San Diego, $^\star$Northeastern University. \\

\textbf{Email:}\\
\texttt{dowu@ucsd.edu, marsgao@uw.edu, xiong.xin@northeastern.edu, m.chinazzi@northeastern.edu, a.vespignani@northeastern.edu, yianma@ucsd.edu, roseyu@eng.ucsd.edu}

\justifying

\textbf{Abstract:}
We introduce DeepGLEAM, a hybrid model for COVID-19 forecasting. DeepGLEAM combines a mechanistic stochastic simulation model GLEAM with deep learning. It uses deep learning to learn the correction terms from GLEAM, which leads to improved performance. We further integrate various uncertainty quantification methods to generate confidence  intervals. We demonstrate DeepGLEAM on real-world COVID-19 mortality forecasting tasks.

\section{Introduction}
Pandemic, hazards, and  disasters  are only increasing in frequency, magnitude and complexity, leading to difficult fiscal, social, cultural, and environmental consequences for the nation and its communities. Accurately modeling of the pandemics dynamics   is the key to our ability to predict the spreading of diseases,  planning for adverse events and reducing disaster losses. However,  the current  machine learning (ML) methods for time series predictions, e.g., COVID-19 case, hospitalizations, and fatalities, still have many limitations.

\textit{Data-driven} methods such as autoregressive models assume stationary distribution. ARIMA \cite{box2015time} model assumes linearity and Gaussian noises, which are too restrictive for the timescales in COVID-19. Deep learning models potentially can alleviate this difficulty but require significant amounts of training data. 
On their own, deep learning (DL) allows flexible modeling of complex, unknown phenomena, but does not integrate physical laws or common sense reasoning. Besides, DL models are often difficult to interpret, and their predictions inaccessible to domain experts \cite{fox2020covid}.

On the other hand, traditional epidemic models are mostly mechanistic models. These models are derived from first principles, can encode the natural history of the disease, and are easy to interpret. For instance, large scale stochastic epidemic models that combine real-world high resolution data on populations and human mobility such as the Global Epidemic and Mobility model (GLEAM)~\cite{balcan2009multiscale,balcan2010modeling,y2018charting,chinazzi2020effect}  characterize complex epidemic dynamics based on meta-population age-structured compartmental models. Due to the difficulty of generating stochastic epidemic profiles for thousands of different subpopulations, these models while providing realistic dynamics of the spatial and temporal spread of an emerging disease, are often slow to simulate at fine resolutions.   Obtaining high-fidelity, high-resolution epidemic models are limited by the high computational cost of generating thousands of multiple embarrassingly parallel model runs to account not only for the uncertainty surrounding the characteristics of the disease outbreak (initial conditions, starting time of the epidemic, etiology of the disease) that might require the exploration of vast parameter space, but also consider the effects of the inherent stochastic behavior of the model. Prior works apply this idea using an ensemble averaging over 50 models for point estimation and uncertainty quantification~\cite{ray2020ensemble,cramer2021evaluation}. Lastly, being the models derived from first principles, they are necessarily making simplifying assumptions and simplifications that can be relaxed by employing a deep learning approach.

We therefore combine both approaches in the current work. We use the forecasts on COVID-19 daily deaths generated by GLEAM~\cite{chinazzi2020effect,davis2020estimating}, an individual-based, stochastic, and spatial epidemic model, to train a diffusion convolution recurrent neural network (DCRNN) that learns stationary features in the time series to improve the alignment between the model predictions and the observed ground-truth data. This combination, named DeepGLEAM, enhances the prediction performance of the stand-alone mechanistic model by allowing to reduce the bias that exists between the model predictions and the observed noisy surveillance data. 

In our experiments, we observe that the DeepGLEAM model significantly outperforms other machine learning models such as the vector auto-regressive (VAR) model and a pure deep learning model in one to four weeks ahead prediction. It also outperforms the GLEAM model in  one to three weeks ahead short term predictions.
We additionally quantify the uncertainty of our predictions.
We compare different uncertainty quantification methods that are commonly considered, including bootstrap, quantile regression, posterior sampling, and approximate Bayesian inference. 
DeepGLEAM demonstrated appealing performances both in accuracy and confidence intervals.

\section{Methodology}

\subsection{Problem Definition}
\begin{figure}[b]
    \centering
    \includegraphics[width=0.45\linewidth]{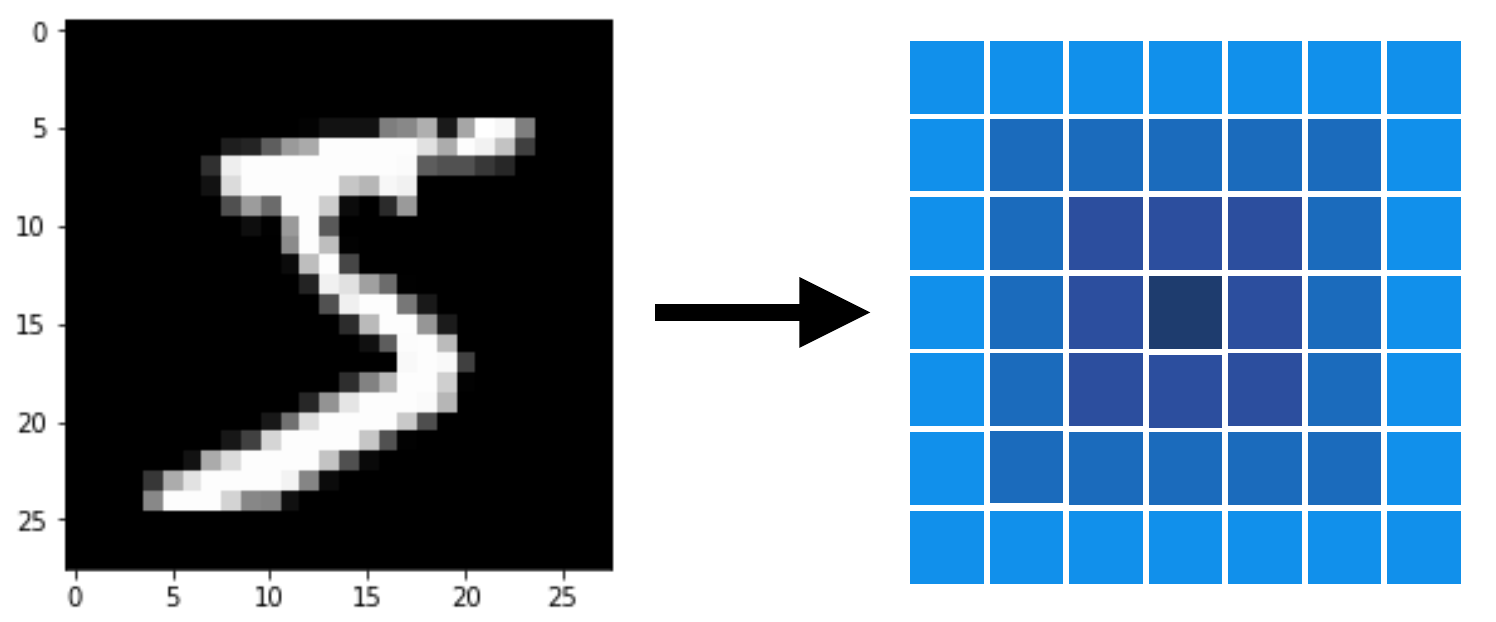}
        \includegraphics[width=0.45\linewidth]{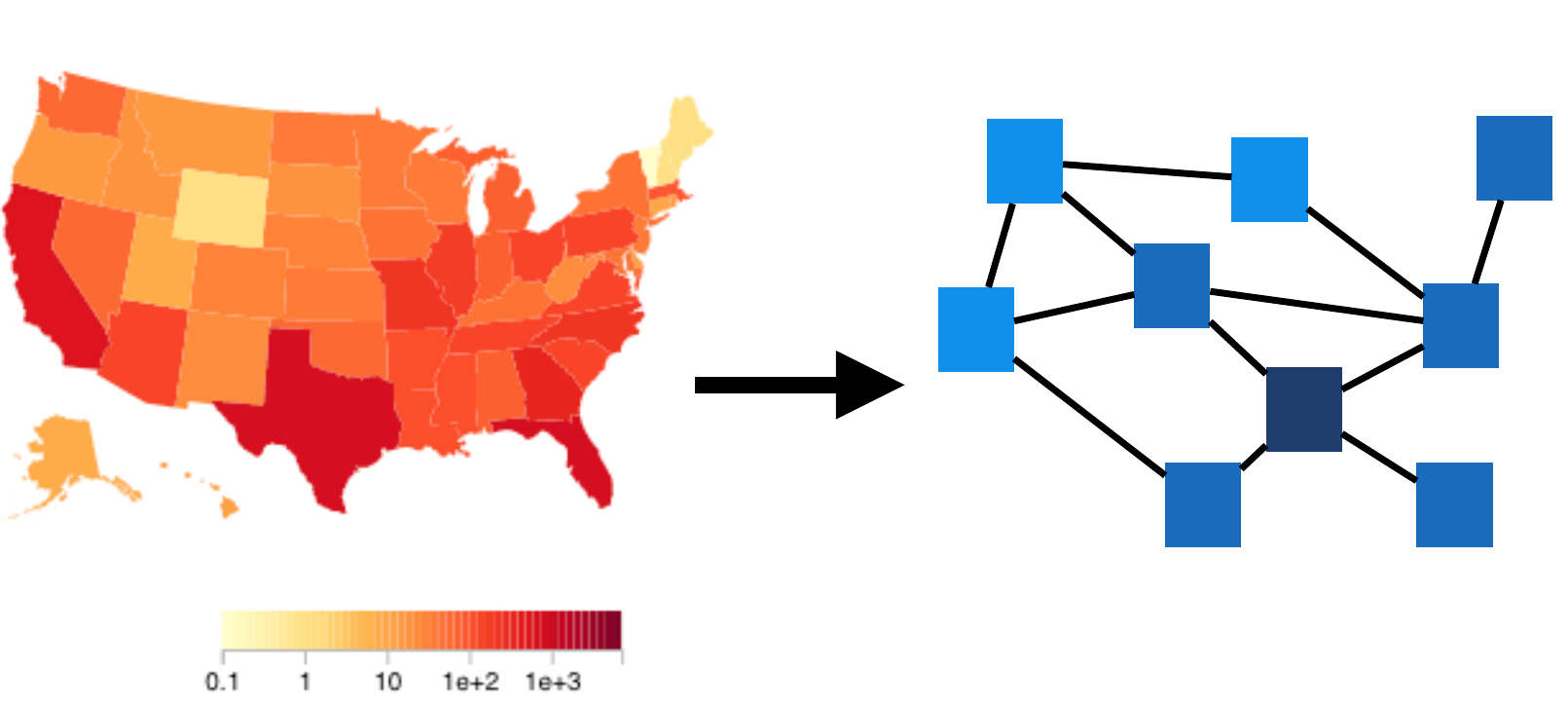}
    \caption{Two types of spatial dependency. left: pixels distributed on the regular grid; right: epidemic diffuses on a irregular grid.   }
    \label{fig:my_label}
\end{figure}

Given multivariate time series $\mathcal{X} = (\V{X}_1, \cdots, \V{X}_t) $ of $t$ time step, with each $\V{X}_t \in \mathbb{R}^{P\times D}$ indicating $D$ features from $P$ locations. In the context of COVID-19 forecasting, locations can be different states or counties. Features can correspond to incident death, and infected cases. In addition to the time series, we also have spatial correlations represented as a graph, corresponding to an adjacency matrix ${A}$.  The spatial correlation can be calculated based on travel data among different locations. Spatiotemporal forecasting aims learns a function $f$ such that:
\begin{equation}
    f: (\mathcal{X}; {A}) \rightarrow (\V{X}_{t+1}, \cdots, \V{X}_{t+H}; {A})
\label{eqn:spatiotemporal_forecast}
\end{equation}
where $H$ is the forecasting horizon. The function $f$ approximate the epidemic dynamics and can either be a mechanistic model or a deep learning model.

\subsection{DeepGLEAM}

\subsubsection{Global Epidemic and Mobility Model.} The Global Epidemic and Mobility model (GLEAM) is a stochastic spatial epidemic model in which the world is divided into over 3,200 geographic subpopulations constructed using a Voronoi tessellation of the Earth's surface. Subpopulations are centered around major transportation hubs (e.g. airports) and consist of cells with a resolution of approximately 25 x 25 kilometers \cite{balcan2009multiscale,balcan2010modeling,tizzoni2012real,zhang2017spread,chinazzi2020effect,davis2020estimating}. High resolution population density data are used to define the number of individuals in each cell \cite{sedac}, while a data-driven approach is employed to generate subpopulation specific age-structured contact patterns \cite{mistry2020inferring}.

GLEAM integrates a human mobility layer - represented as a network - that uses both short-range (i.e. commuting) and long-range (i.e. flights) mobility data from the Offices of Statistics for $30$ countries on $5$ continents as well as the Official Aviation Guide (OAG) and IATA databases (updated in $2020$). The air travel network consists of the daily passenger flows between airport pairs (origin and destination) worldwide mapped to the corresponding subpopulations. Where information is not available, the short-range mobility layer is generated synthetically by relying on the “gravity law”  or the more recent “radiation law”  both calibrated using real data \cite{Simini2012}.  

The model is calibrated to realistically describe the evolution of the COVID-19 pandemic as detailed in \cite{chinazzi2020effect,davis2020estimating}.
Lastly, GLEAM is stochastic and produces an ensemble of possible epidemic outcomes for each set of initial conditions. To account for the potentially different reporting levels of the states, a free parameter Infection Fatality Rate (IFR) multiplier is added to each model. To calibrate and select the most reasonable outcomes, we filter the models by the latest hospitalization trends and confirmed cases trends, and then we select and weight the filtered models using Akaike Information Criterion \cite{zhang2017forecasting}. The forecast of the evolution of the epidemic is formed by the final ensemble of the selected models.

\begin{figure}    \centering
    \includegraphics[scale=0.35, trim = 0 10 0 80]{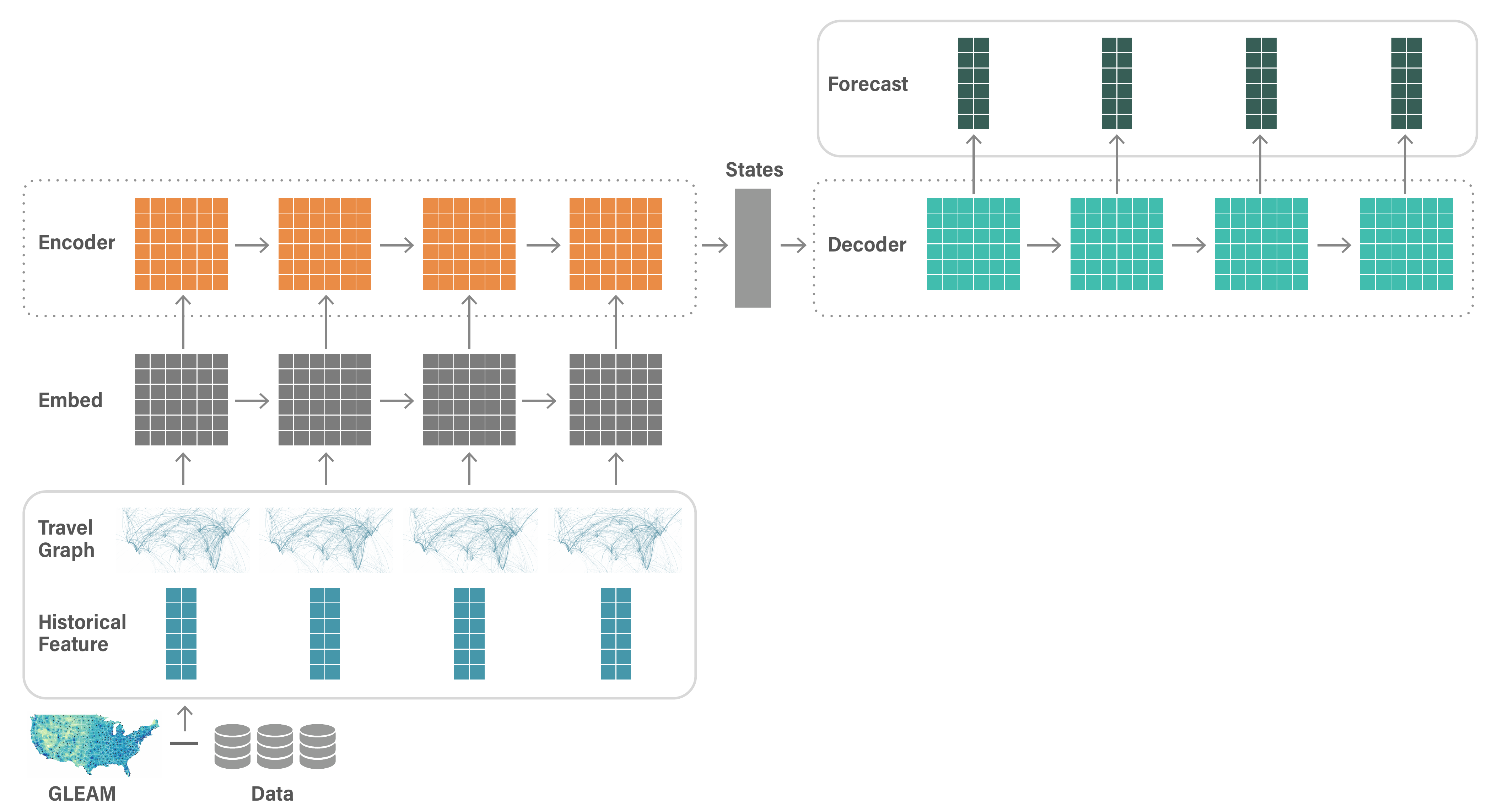}
    \caption{System architecture of DeepGLEAM. A deep learning model (DCRNN) is trained with the residual between reported incident death and predictions from GLEAM.}
    \label{fig:gleam}
\vspace{5mm}
\end{figure}

\subsubsection{Deep Learning Model: Diffusion Convolutional RNN}
When the locations are evenly spaced on a regular grid, popular deep forecasting models   include Convolutional LSTM  (ConvLSTM)~\cite{xingjian2015convolutional} and PredRNN~\cite{wang2017predrnn}. The dynamics is approximated by a recurrent neural network (RNN), 
\begin{equation}
    \V{h}_{t+1}= \sigma(W^h\cdot \V{h}_{t} + W^x\cdot \V{X}_{t}) 
\label{eqn:convlstm}
\end{equation}
where $\V{h}_t $ are the time-varying hidden states. The convolution operator extracts  spatial features and uses an RNN to model the temporal dynamics, which is the key idea behind ConvLSTM \cite{xingjian2015convolutional}.

When the locations are distributed as a graph which is the case for COVID-19 forecasting, a natural extension is to generalize  convolution to graph convolution.  This type of models include Spatiotemporal Graph CNN~\cite{yu2018spatio}, Diffusion Convolutional RNN (DCRNN)~\cite{li2018diffusion} and other variants.

Mathematically speaking, that means to replace the matrix multiplication operator in an RNN with a  graph convolution operator, leading to the following design of DCRNN \cite{li2018diffusion}:
\begin{equation}
   \V{h}_{t+1}= \sigma(W \ast_g \V{h}_{t} + W\ast_g \V{X}_{t})
\label{eqn:graph_convlstm}
\end{equation}
Where $\ast_g$ stands for graph convolution. There are many variations for graph convolution or graph embedding. One example is  
\begin{equation}
W \ast_g \V{X}_t = W\cdot (D^{-1}A)\cdot \V{X}_t
\label{eqn:convolution}
\end{equation}
Here $D \in \mathbb{R}^{P\times P}$ contains the diagonal element of $A$.  One can also replace the random walk matrix $D^{-1}A$ with the normalized Laplacian matrix $I -D^{1/2}(D-A)D^{1/2}$ as in \cite{kipf2016semi}.
The entire network is trained by maximizing the likelihood of generating the target future time series using backpropagation through time. DCRNN is able to capture spatiotemporal dependencies in time series without imposing strong modeling assumptions on the dynamics.

\begin{figure}[tbp]
    \centering
        \includegraphics[width=.9\linewidth]{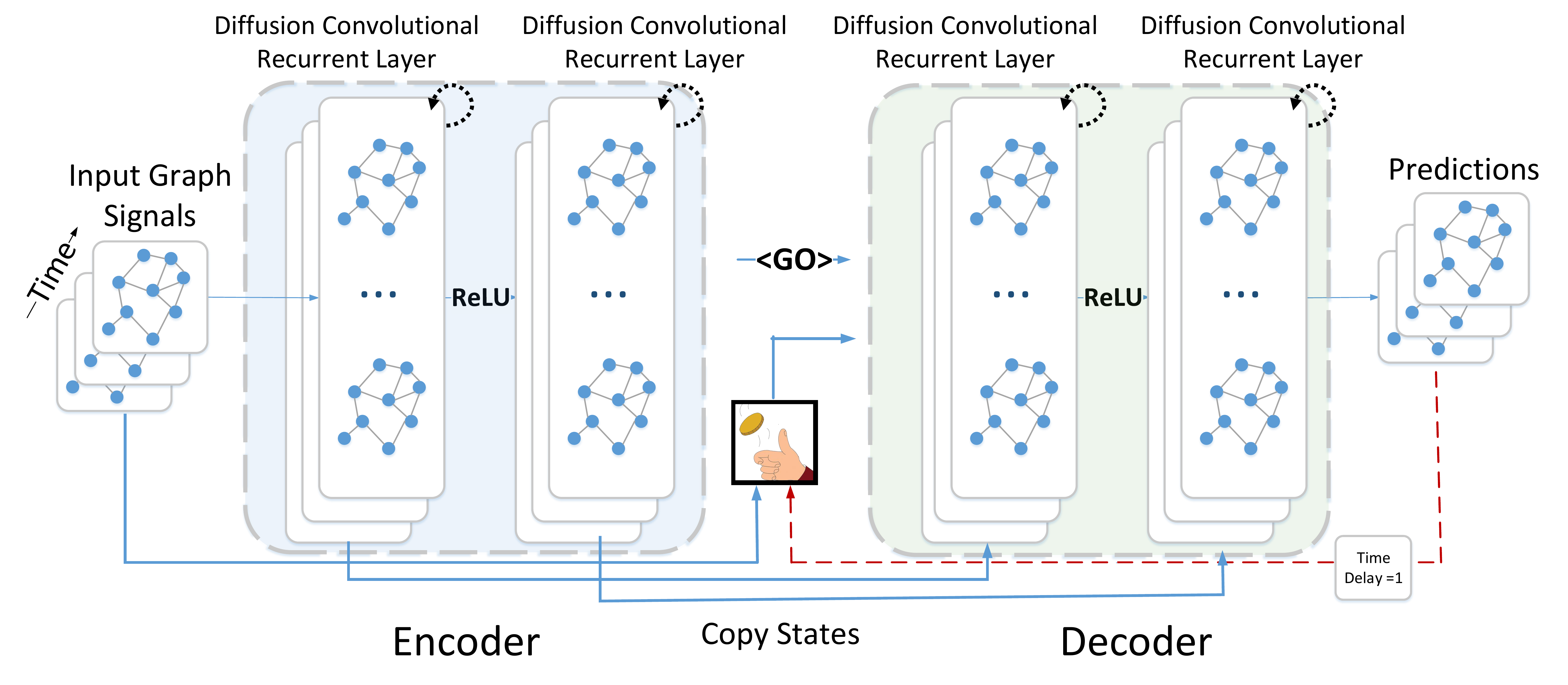}
    \caption{System architecture for the \textit{Diffusion Convolutional Recurrent Neural Network} designed for spatiotemporal traffic forecasting. The historical time series are fed into an encoder whose final states are used to initialize the decoder. The decoder makes predictions based on either previous ground truth or the model output.}
    \label{fig:system_architecture}
\end{figure}
\subsubsection{DeepGLEAM: combining GLEAM with Deep Learning} 
We use the residual between daily death number and GLEAM predictions as input and output of the DCRNN model and name it DeepGLEAM model. 
We use an encoder-decoder sequence to sequence learning framework in the DCRNN structure.
The encoder reads as input a $7 \times 50 \times 4$ tensor that comprises the daily residuals between the observed death number and the GLEAM forecasts for the $50$ US states over $4$ different prediction horizons.
It encodes the information in $7$ hidden layers. 
The decoder produces forecasts for each state for the following $4$ weeks.
We perform autoregressive weekly death predictions (from one week ahead to four weeks ahead).  
To this end, we build a hybrid DeepGLEAM model. The model  architecture of DeepGLEAM is shown in Figure ~\ref{fig:gleam}. 
The default DeepGLEAM model is for point estimation. Next, we discuss probabilistic forecasts by combing DeepGLEAM with  Frequentist or Bayesian uncertainty quantification (UQ) methods, including bootstrap, quantile regression (Quantile), spline quantile regression (SQ), MIS regression (MAE-MIS), MC dropout, and SG-MCMC.

\subsection{Uncertainty Quantification}
In this subsection, we introduce various methods for uncertainty quantification that can be seamlessly integrated into DeepGLEAM. 

\paragraph{Bootstrap.}
The (generalized) bootstrap method~\cite{efron2016} randomly generates in each round a weight vector over the index set of the data.
The data are then resampled according to the weight vector.
With every resampled dataset, we retrain our model and generate predictions.
Using the predictions from different retrained models, we inference the $(1-\rho)$ confidence interval using $(1-\frac{\rho}{2})$ and $\frac{\rho}{2}$ quantiles. The quantiles can be estimated using the order statistics of Monte Carlo samples.

\paragraph{Mean Interval Score (MIS) and Quantile Regression}
For a fixed confidence level $\rho$, we can directly minimize Mean Interval Score (MIS) to obtain estimates of the confidence intervals. MIS is defined for evaluation of estimated upper and lower confidence bounds.
For a random vector $Z\sim\mathbb{P}_Z$, if the estimated upper and lower confidence bounds are $u$ and $l$, where $u$ and $l$ are the $(1-\frac{\rho}{2})$ and $\frac{\rho}{2}$ quantiles for the $(1-\rho)$ confidence interval, MIS is defined using samples $z_i\sim\mathbb{P}_Z$
\begin{equation}
\label{eqn:mis_def}
MIS_N(u,l) = \frac{1}{N} \sum_{i=1}^N \lrp{ \lrp{u-l} + \frac{2}{\rho}(z_i-u)\ind{z_i>u} + \frac{2}{\rho}(l-z_i)\ind{z_i<l} }.
\end{equation}

Specifically, to use MIS as a loss function for deep neural networks, we use a multi-headed model to jointly output the upper bound $u(x)$, lower bound $l(x)$, and the  prediction $f(x)$ for a given input $x$, and minimize the neural network parameter $\theta$:
\begin{equation}
\begin{split}
L_{\text{MIS}}(y,u(x),l(x),f(x);\theta, \rho) =& \min\limits_{\theta}\Big\{ \mathbb{E}_{(x,y)\sim \mathcal{D}}[ (u(x)-l(x))+ \frac{2}{\rho}(y-u(x))\ind{y>u(x)}\\ 
& + \frac{2}{\rho}(l(x)-y)\ind{y<l(x)} + |y-f(x)|]   \Big\} \label{eqn:MIS_loss}
\end{split}
\end{equation}

Here $\ind{}$ is an indicator function, which can be implemented using the identity operator over the larger element in Pytorch.

For quantile regression~\cite{Koenker1978,Koenker2005}, we can use the one-sided quantile loss function to generate predictions for a fixed confidence level  $\rho$. Given an input $x$, and the output $f(x)$ of a neural network, parameterized by $\theta$, quantile loss is defined as follows:
\begin{equation}
L_{\text{Quantile}}(y, f(x); \theta, \rho) =\min\limits_{\theta}\Big \{ \mathbb{E}_{(x,y)\sim \mathcal{D}}[(y-f(x))(\rho-\ind{y<f(x)}]\Big \}
 \label{eqn:pinball_loss}
\end{equation}

Quantile regression  behaves similarly as the MIS regression method. Both methods generate one confidence interval per time. In addition, \cite{kivaranovic2020adaptive,  tagasovska2019single,pearce2018high} have explored  similar ideas of directly optimizing the prediction interval using different variations of quantile loss.

One caveat of these methods is that different predicted quantiles can cross each other due to variations given finite data.
This will cause a strange phenomenon when the size of the data set and the model capacity is limited: the  higher confidence interval does not contain the interval of lower confidence level or even the point estimate.
One remedy for this issue is to add variations in both the data and the parameters to increase the effective data size and model capacity.
In particular, during training, we can use different subsets of data and repeat random initialization from a prior distribution to form an ensemble of models.
In this way, our modified MIS and quantile regression methods have integrated across different models to quantify the prediction uncertainty and have taken advantage of the Bayesian philosophies.

Another solution to alleviate quantile crossing  and unify different confidence levels is to minimize CRPS by assuming the quantile function to be a piecewise linear spline with
monotonicity \cite{gasthaus2019probabilistic}, a method we call Spline Quantile regression (SQ).
In the experiment, we also included this method for comparison.

\paragraph{Stochastic Gradient MCMC (SG-MCMC)}
In Bayesian's perspective, the posterior distribution contains belief about the parameter based on observed data. Therefore, we could construct confidence interval via samplings from posterior distribution. To estimate expectations or quantiles according to the posterior distribution over the parameter space, we use SG-MCMC~\cite{welling2011bayesian,ma2015complete}.
We find in the experiments that the stochastic gradient thermostat method (SGNHT)~\cite{ding2014bayesian,shang2015} is particularly useful in controlling the stochastic gradient noise.
This is consistent with the observation in~\cite{heek2019} where stochastic gradient thermostat method is applied to an i.i.d. image classification task.

\paragraph{Approximate Bayesian Inference}
SG-MCMC can be computationally expensive. There are also approximate Bayesian inference methods introduced to accelerate the inference procedures~\cite{maddox2019simple,dusenberry2020}.
In particular, the Monte Carlo (MC) drop out method sets some of the  network weights to zero according to a prior distribution~\cite{gal2016dropout,gal2017concrete}.
This method serves as a simple alternative to  variational Bayes methods which approximate the posterior~\cite{blundell2015weight,graves2011practical,louizos2016structured,rezende2014stochastic,qiu2019quantifying}.
We examine the popular MC dropout method in the experiments for comparison. 
\section{Experiments}
We evaluate the performance of both Frequentist and Bayesian UQ methods on the time-series dataset for COVID-19 incident deaths. 
This is highly challenging as the data is very small, highly noisy, and pertain complex spatial dependency.  The experiments are implemented using pytorch \cite{paszke2019pytorch}. Based on our observations, Bayesian methods typically reach lower error in their mean predictions while Frequentist methods, especially quantile and MIS regression, prevail in providing reasonable confidence intervals and achieve the best performance in MIS.

\subsection{Datasets and Experiment Setup}
\paragraph{Datasets.}

The COVID-19  dataset  contains reported death from Johns Hopkins University \cite{dong2020interactive} and the death predictions from a mechanistic, stochastic, and spatial metapopulation epidemic model called Global Epidemic and Mobility Model (GLEAM) \cite{balcan2009multiscale, balcan2010modeling, tizzoni2012real, zhang2017spread, chinazzi2020effect}. Both data are recorded for the $50$ US states during the time period from May 24th to Sep 12th 2020. We use the residual between the reported death and the corresponding GLEAM predictions to train the model (\url{http://covid19.gleamproject.org/}).

We form the spatial graph---adjacency matrix $A$ in Eqn~\eqref{eqn:spatiotemporal_forecast}---using air travel flows between different states.
The mobility data is obtained from the Official Aviation Guide (OAG) and the International Air Transportation Association (IATA) databases (updated in $2020$). Each directed weighted edge of the graph represents the average number of passengers traveling between two states on a daily basis.

\begin{figure}    \centering
    \includegraphics[scale=0.6, trim = 0 10 0 80]{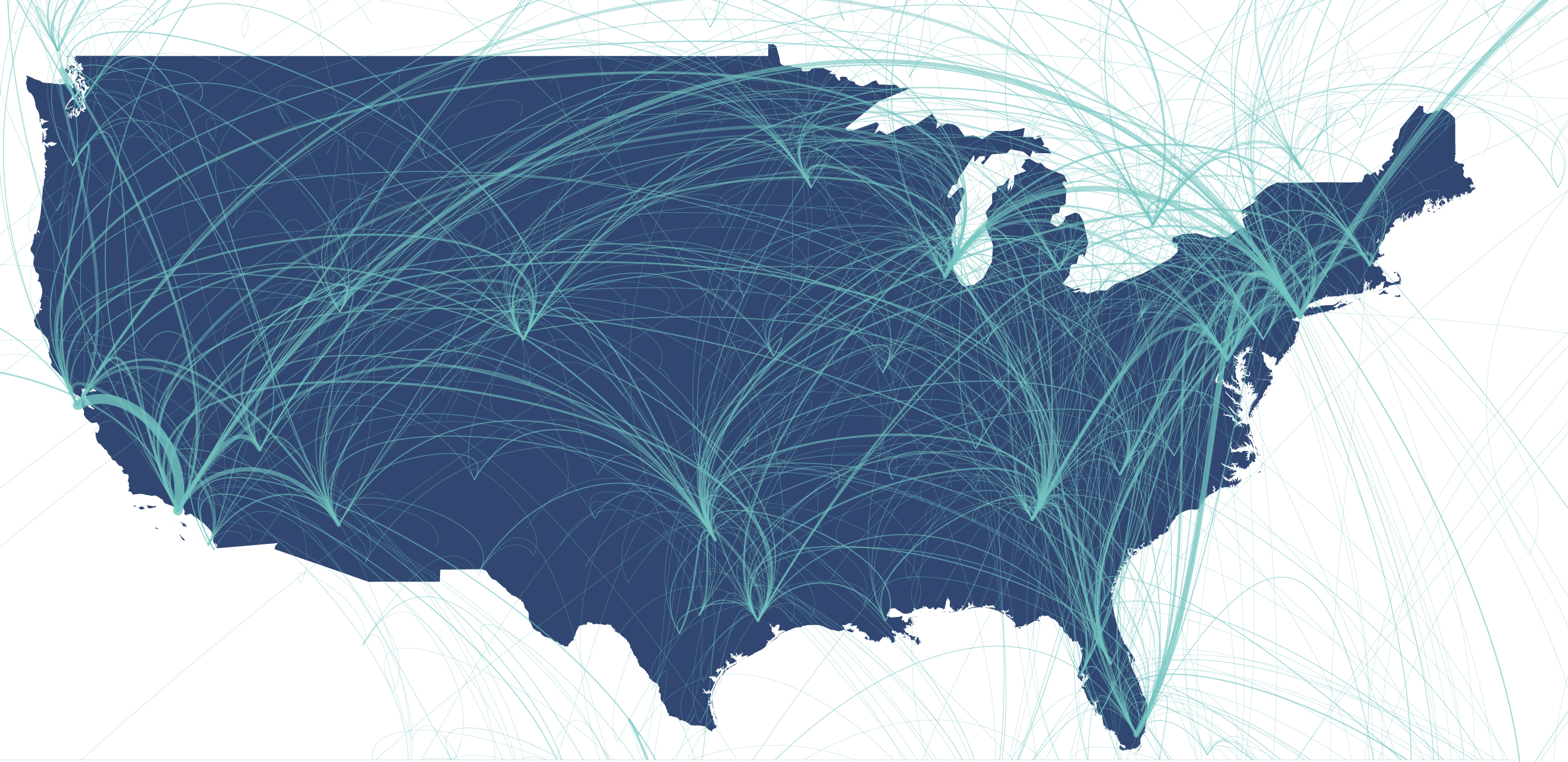}
    \caption{Original flight network connecting US airports. Data is aggregated at the State level to construct the network State-to-State graph.}
    \label{fig:flight}
\vspace{5mm}
\end{figure}

\paragraph{Evaluation Metrics.}
We evaluate our model using the metrics. For point estimation, we define our metrics using Mean Absolute Error (MAE) and Root Mean Square Error (RMSE). Given multivariate time series $\mathcal{X}_{[1,t]} = (\V{X}_1, \cdots, \V{X}_t) $ of $t$ time step, with each $\V{X}_t \in \mathbb{R}^{P\times D}$ indicating $D$ features from $P$ locations.
Suppose $\mathcal{X}_{[t+1, t+h]}=\{\V{X}_{t+1}, \dots, \V{X}_{t+h}\}$ is the ground truth data and $\hat{\mathcal{X}}_{[t+1, t+h]}=\{\hat{\V{X}}_{t+1}, \dots, \hat{\V{X}}_{t+h}\}$ is the prediction. To evaluate how well the prediction $\hat{\mathcal{X}}$ is, we could define our metrics as below: \\
\begin{equation}
MAE(\mathcal{X}_{[t+1, t+h]}, \hat{\mathcal{X}}_{[t+1, t+h]})=\frac{1}{D\cdot P\cdot h}\sum_{i=1}^h ||\mathcal{X}_{[t+1, t+h]}-\hat{\mathcal{X}}_{[t+1, t+h]} ||_{1,1}.
\end{equation}

\begin{equation}
RMSE(\mathcal{X}_{[t+1, t+h]}, \hat{\mathcal{X}}_{[t+1, t+h]})=\frac{1}{D\cdot P\cdot h}\sum_{i=1}^h ||\mathcal{X}_{[t+1, t+h]}-\hat{\mathcal{X}}_{[t+1, t+h]} ||_{2,2}^2,
\end{equation}
where is a matrix norm defined by $||A||_{m,n}=\lrp{\sum_{i=1}^D \lrp{\sum_{i=1}^P |a_{ij}^m|}^{\frac{n}{m}}}^{\frac{1}{n}}$ .

For uncertainty quantification, we define our metric via MIS as defined in \eqref{eqn:mis_def}. MIS rewards narrower confidence or credible intervals and penalizes intervals that do not include the observations. In our case, MIS is preferred  over other scoring functions such as the Brier score \cite{brier1950verification}, Continuous Ranked Probability Score (CRPS) \cite{matheson1976scoring, hamill1995probabilistic, gneiting2007strictly} as it is intuitive and  easy to compute.

\paragraph{Experiment Setup for DCRNN in various contexts.} 

For point estimation using DCRNN, we apply  one hidden layer of 8 RNN units and Laplacian filters. The optimization process is via Adam optimizer \cite{kingma2014adam} learning rate $1e^{-2}$.

For frequentist uncertainty quantification techniques, Bootstrap method specifically is leave-one-out bootstrap due to the limited sample size. For Quantile regression,
we apply pinball loss function \cite{Koenker1978,Koenker2005} to train three different quantiles for the corresponding confidence interval (CI), e.g., (0.025, 0.5, 0.975) for 95\%. For SQ regression,
we use spline quantile function to approximate a quantile function and use the CRPS as the loss function. For MIS regression, 
We combine MAE with MIS and directly minimize this loss function.

For Bayesian uncertainty quantification methods, considering MC Dropout, we apply random dropout through the testing process with 5\% drop rate and iterate 300 times to achieve a stable prediction. The result of SG-MCMC is averaged from 25 posterior samples, and we selected a Gaussian prior $\mathcal{N}(0, 2.0)$ with random initialization as $\mathcal{N}(0, 0.05)$. 

\subsection{Performance of DeepGLEAM and other baseline methods } 

To validate the benefit of hybrid modeling, we  compare with the pure deep learning model (Deep) for point estimation, the pure mechanistic GLEAM model, the DeepGLEAM model, and the SG-MCMC model. To justify the use of deep learning, we also compare with the classic time series model: Vector Auto-regressive (VAR) model.  
In Table \ref{tab:deep}, we can see that  VAR  clearly under-performs other methods. In general, shallow models (VAR) fails in this task due to the complex dynamic of COVID-19 prediction.

\begin{figure}[t!]
    \centering
    \includegraphics[scale=0.35, trim = 0 35 0 10 ]{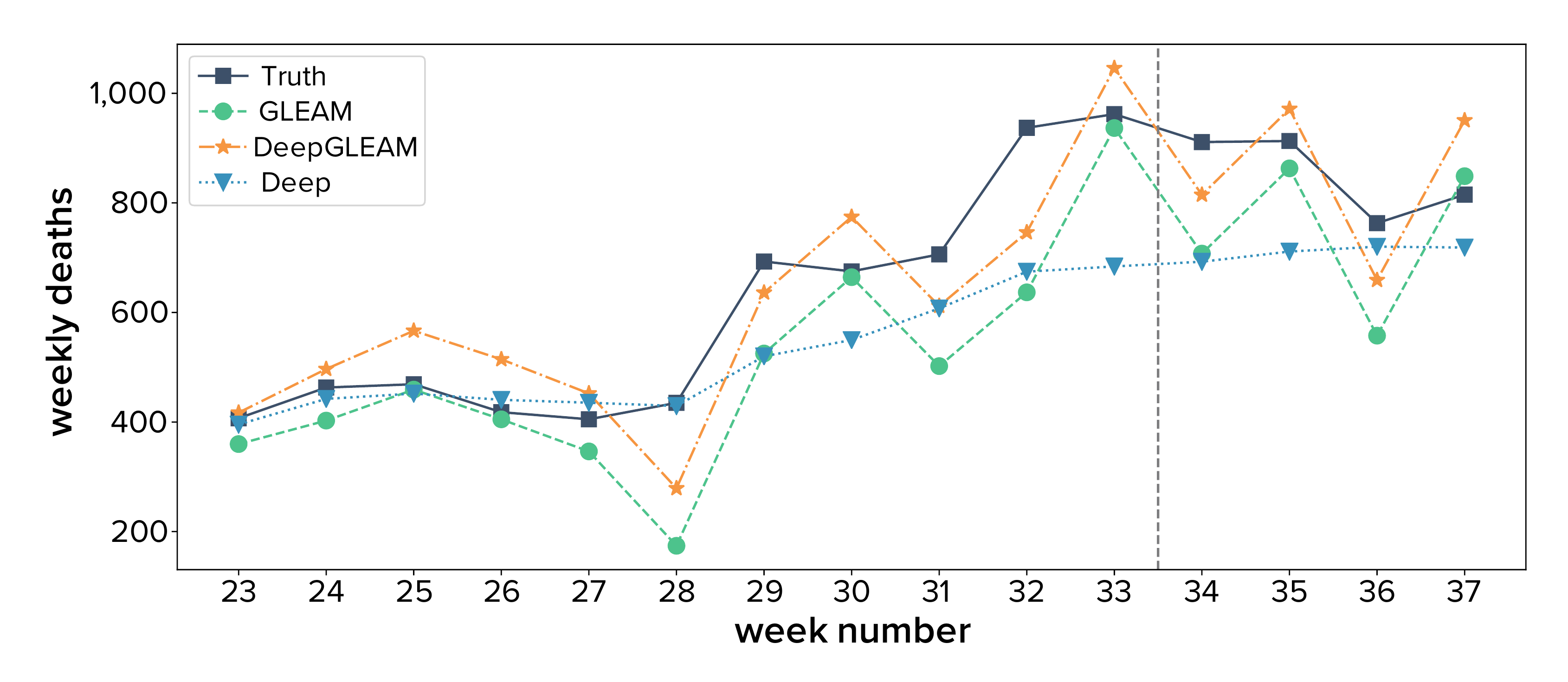}
    \caption{One week ahead COVID-19 forecasts visualization in California. Comparison shown for GEAM, DeepGLEAM and Deep models.}
    \label{fig:dcrnn_vs_deepgleam}
\end{figure}
\begin{table}[t!]
\centering
\begin{tabular}{c|ccccc}
\toprule
Horizon $H$ & VAR & Deep & GLEAM & DeepGLEAM & SG-MCMC\\ \hline
{1W} & 159.29 & 239.94 & 73.59 & 66.03 & 58.30\\
\cline{1-6}
{2W}& 305.99 & 213.31 & 65.46& 57.67 & 46.61\\
\cline{1-6}
{3W}& 456.35 & 189.49 & 73.75& 70.12 & 59.27 \\
\cline{1-6}
{4W}& 697.27 & 161.88 & 70.16 & 70.75 & 70.57\\
\bottomrule
\end{tabular}
\caption{RMSE comparison of different approaches for COVID-19 mortality forecasting}
\label{tab:deep}
\end{table}

In Figure \ref{fig:dcrnn_vs_deepgleam}, we visualize an example in California to compare the forecasts of different models. The input data (for training or validation) are from weeks before $34$, and we start the prediction from week $34$, labeled by the vertical dash line. It can be observed that Deep model fails to predict the dynamics of COVID-19 evolution. We quantitively compare the accuracy among DeepGLEAM, SG-MCMC, Deep, and GLEAM models using RMSE. DeepGLEAM and Deep are deterministic models while SG-MCMC is the statistical model with the best accuracy. Table \ref{tab:deep} shows both DeepGLEAM and SG-MCMC outperform GLEAM while the Deep model is much worse than GLEAM. By calculation, there is a $6.6\%$ improvement for DeepGLEAM and $17\%$ improvement for SG-MCMC on average from GLEAM.

One reason for this phenomenon is the distribution shifts in the epidemiology dynamics.
Without background knowledge, deep neural networks do not have enough inductive bias to guide their predictions.
On the other hand, GLEAM, DeepGLEAM, and SG-MCMC leverage the mechanistic knowledge about disease transmission dynamics to infer underlying dynamic change.

\subsection{Comparison of uncertainty quantification methods}

Table \ref{tab:covid_com} compares 6 different UQ methods for 1-4 weeks ahead forecasting of COVID-19 mortality.  We observe that SG-MCMC performs the best in RMSE and MAE.  Quantile and MIS regressions achieve the best performance in MIS.
SQ regression suffers from a poor MIS due to the small number of knots for linear spline quantile function construction. For MC dropout, its prediction accuracy is relatively robust but its MIS is bad. The interval in Table \ref{tab:covid_com} shows  Bootstrap, SQ, MC dropout, and SG-MCMC methods tend to make overconfident predictions with shorter intervals and larger coverage compared with Quantile, and MIS methods.\

\begin{figure}[h]
    \centering
    \includegraphics[scale=0.24, trim = 20 35 40 10 ]{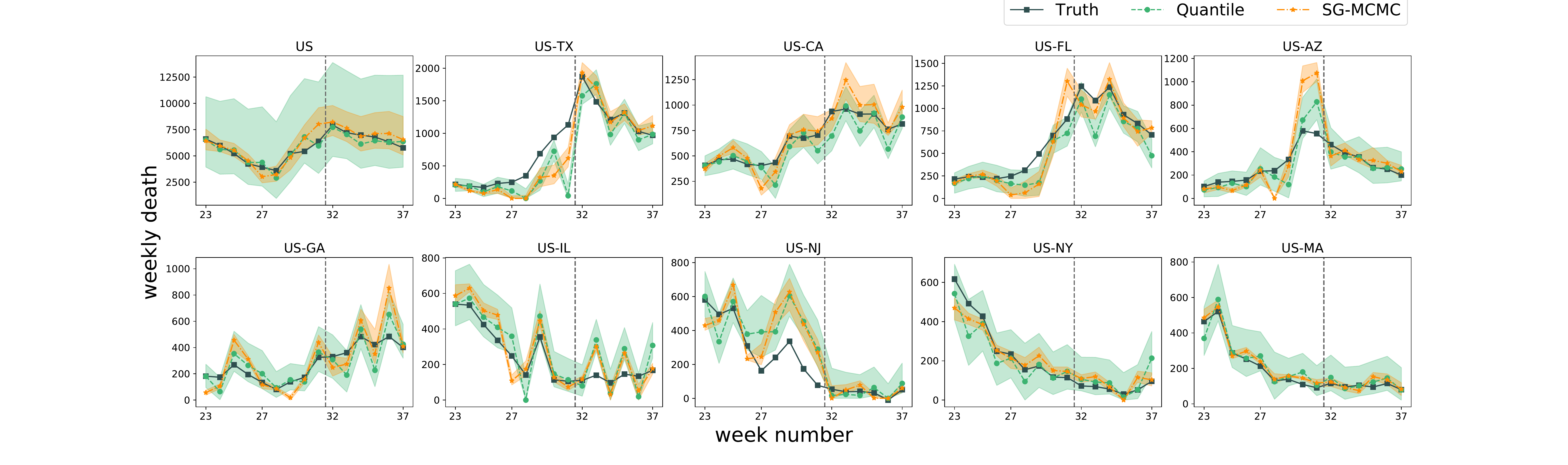}
    \caption{One week ahead COVID-19 prediction in the country level and 9 states with largest death from week 23 to week 37.}
    \label{fig:covid}
    \vspace{-3mm}
\end{figure}

We visualize the predictions of quantile regression and SG-MCMC in Figure \ref{fig:covid}. We find mean predictions from SG-MCMC are closer to the ground truth while quantile regression provides better confidence bounds at the state level. For example, the US-TX subplot shows that the SG-MCMC's mean prediction is closer to the ground truth compared with quantile prediction. Meanwhile, the US-GA and US-NY subplots show that the overconfident credible interval of SG-MCMC fails to cover the ground truth. In these cases, quantile regression can provide safer confidence bounds and achieve better MIS. 
For country-level prediction, as shown in the US subplot, SG-MCMC outperforms quantile regression in both mean prediction accuracy and confidence bounds.

\begin{table}[t!]
\centering
\resizebox{\columnwidth}{!}{
\begin{tabular}{c|c|cccccc}
\toprule
Horizon $H$ & Metric & Bootstrap & Quantile & SQ & MIS & MC Dropout & SG-MCMC\\ 
\midrule
\multirow{2}{*}{1W} & RMSE & 64.63 & 70.42 & 71.72 & 74.07 & 66.82 & \textbf{58.30} \\ 
& MAE & 32.57 & 36.63 & 34.42 & 39.03 & {34.27} & {\textbf{29.72}}  \\ 
& MIS (95$\%$ CI) & 856.87 & \textbf{413.37} & 1049.69 & 427.53 & 790.24 & 563.77\\ 
& {Interval} & {32.48} & {190.98} & {23.73} & {227.19} & {47.79} & {47.05}\\ 
& {MIS (90$\%$ CI)} & {444.68} & {302.77} & {541.61} & {\textbf{285.99}} & {443.13} & {365.6}\\
& {Interval} & {32.48} & {129.92} & {22.62} & {141.24} & {40.51} & {43.04}\\ 
& {MIS (80$\%$ CI)} & {252.39} & {218.73} & {286.35} & {\textbf{200.32}} & {254.42} & {212.97}\\
& {Interval} & {24.28} & {73.23} & {20.31} & {84.45} & {31.94} & {35.96}\\ 
\midrule
\multirow{2}{*}{2W}& RMSE & 54.21 & 61.35 & 63.32 & 64.42 & 57.63 & \textbf{46.61}\\ 
& {MAE} & {32.38} & {36.81} & {34.03} & {36.35} & {33.64} & {\textbf{27.65}}  \\ 
& MIS (95$\%$ CI) & 762.55 & \textbf{363.14} & 1010.97 & 379.27 & 686.43 & 599.14\\ 
& {Interval} & {36.25} & {219.06} & {24.46} & {260.24} & {55.93} & {45.45}\\ 
& {MIS (90$\%$ CI)} & {399.40} & {276.32} & {522.99} & {\textbf{270.41}} & {397.52} & {332.00}\\
& {Interval} & {36.25} & {150.54} & {23.35} & {161.9} & {47.53} & {42.19}\\ 
& {MIS (80$\%$ CI)} & {235.51} & {196.56} & {277.86} & {\textbf{185.53}} & {236.7} & {197.59}\\
& {Interval} & {27.69} & {85.69} & {21.00} & {97.11} & {37.58} & {35.75}\\ 
\midrule
\multirow{2}{*}{3W}& RMSE & 67.70 & 72.92 & 72.52 & 73.65 & 70.03 & \textbf{59.27}\\ 
& {MAE} & {40.33} & {44.29} & {41.24} & {43.15} & {41.26} & {\textbf{34.62}}  \\ 
& MIS (95$\%$ CI) & 1028.63  & 411.05 & 1292.98 & \textbf{402.46} & 905.66 & 821.71\\ 
& {Interval} & {39.95} & {242.47} & {24.16} & {291.96} & {62.39} & {46.16}\\ 
& {MIS (90$\%$ CI)} & {534.29} & {315.96} & {664.50} & {\textbf{304.12}} & {515.15} & {443.94}\\
& {Interval} & {39.95} & {170.12} & {23.09} & {184.03} & {53.10} & {43.07}\\ 
& {MIS (80$\%$ CI)} & {307.14} & {237.90} & {349.00} & {\textbf{220.60}} & {300.09} & {254.54}\\
& {Interval} & {29.87} & {96.07} & {20.81} & {111.79} & {42.03} & {36.63}\\ 
\midrule
\multirow{2}{*}{4W}& RMSE & 68.63 & 73.92 & \textbf{69.94} & 72.44 & 70.60 & 70.57\\ 
& {MAE} & {41.71} & {46.20} & {41.79} & {44.45} & {42.28} & {\textbf{40.66}}  \\ 
& MIS (95$\%$ CI) &  1035.26& 455.27 & 1303.02 & \textbf{428.82} & 891.45 & 852.26\\ 
& {Interval} & {43.61} & {262.09} & {23.85} & {316.13} & {67.52} & {47.58}\\ 
& {MIS (90$\%$ CI)} & {539.43}  & {359.69} & {669.76} & {\textbf{343.83}} & {512.72} & {458.94}\\
& {Interval} & {43.61} & {190.23} & {22.79} & {206.27} & {57.50} & {44.65}\\ 
& {MIS (80$\%$ CI)} & {315.05} & {262.76} & {351.96} & {\textbf{252.51}} & {302.66} & {261.32}\\
& {Interval} & {32.32} & {105.6} & {20.58} & {128.51} & {45.56} & {38.03}\\ 
\bottomrule
\end{tabular}}
\caption{Performance comparison of different approaches on Autoregressive DeepGLEAM Model for COVID-19 mortality forecasting. }
\label{tab:covid_com}
\end{table}

\subsection{Predictions of the rest 41 states in the U.S.}
\begin{figure}[t!]
    \vspace{5mm}
    \centering
    \includegraphics[scale=0.12, trim = 0 35 0 120 ]{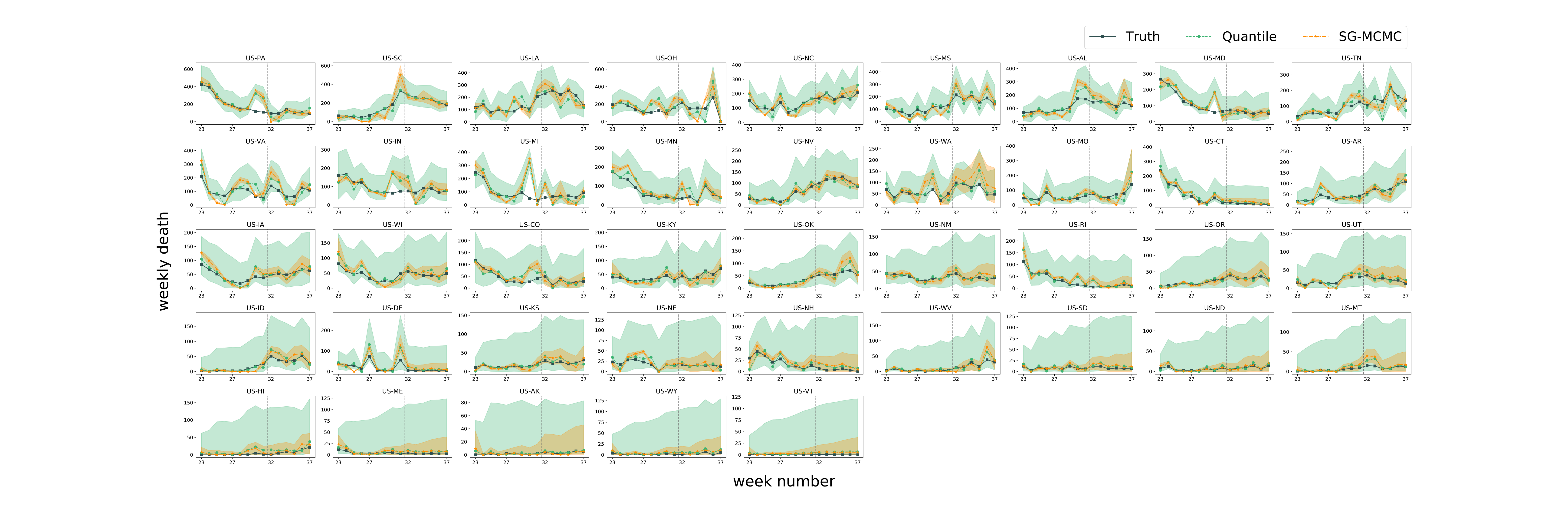}
    \caption{One week ahead COV-19 prediction for the rest 41 states. }
    \label{fig:covid_rest}
\end{figure}

Figure \ref{fig:covid_rest} shows the predictions of the rest 41 states in the U.S.. The data before week 32 has been seen during training or evaluation, therefore we focus on the result to the right of the vertical dashed line.

\newpage

\section{Conclusion}
In this paper, we introduce DeepGLEAM, a hybrid model for COVID-19 forecasting. DeepGLEAM combines a mechanistic stochastic simulation model, GLEAM, with deep learning. It uses deep learning to learn the correction terms from GLEAM, which leads to improved performance. By integrating various uncertainty quantification methods, we demonstrate DeepGLEAM on real-world COVID-19 mortality forecasting tasks in both point estimation and uncertainty quantification. The experimental results show that DeepGLEAM outperforms baseline methods including Vector Autoregressive model, deep learning model, and GLEAM respectively.

\pagebreak

\bibliographystyle{plain}
\bibliography{ref}

\newpage
\appendix
\section{Comparison among methods}
 \begin{table}[h]
  \centering
  \resizebox{\columnwidth}{!}{
\begin{tabular}{cccc|cccc}
    \toprule                
    Method & Parallel computing resource & Small sample & Asymptotic consistency  &  & Accuracy & Uncertainty  \\
    \midrule
 \texttt{Bootstrap} & 25 &  & \checkmark &  & \checkmark & \\
 \texttt{Quantile} & 1 & & &  & \checkmark & \checkmark\\
 \texttt{MIS} & 1 & & &  & \checkmark & \checkmark\checkmark \\
 \texttt{MC Dropout} & 1 & & &  & \checkmark & \\
 \texttt{SG-MCMC} & 25 & \checkmark & \checkmark & & \checkmark\checkmark & \checkmark \\
    \bottomrule
  \end{tabular}
  }
    \caption{Comparison of different deep uncertainty quantification methods for forecasts. Double check marks represent robustly highest performance in experiments. 
 }
 \label{tb:comparison-table}
\end{table}

\section{COVID-19 forecasting experiment}
\subsection{Global Epidemic and Mobility Model.} \label{app:covid_deepgleam}

The Global Epidemic and Mobility model (GLEAM) is a stochastic spatial epidemic model in which the world is divided into over 3,200 geographic subpopulations constructed using a Voronoi tessellation of the Earth's surface. Subpopulations are centered around major transportation hubs (e.g. airports) and consist of cells with a resolution of approximately 25 x 25 kilometers \cite{balcan2009multiscale,balcan2010modeling,tizzoni2012real,zhang2017spread,chinazzi2020effect,davis2020estimating}. High resolution data are used to define the population of each cell \cite{sedac}. Other attributes of individual subpopulations, such as age-specific contact patterns, health infrastructure, etc., are added according to available data \cite{mistry2020inferring}.

GLEAM integrates a human mobility layer - represented as a network - that uses both short-range (i.e. commuting) and long-range (i.e. flights) mobility data from the Offices of Statistics for $30$ countries on $5$ continents as well as the Official Aviation Guide (OAG) and IATA databases (updated in $2020$). The air travel network consists of the daily passenger flows between airport pairs (origin and destination) worldwide mapped to the corresponding subpopulations. Where information is not available, the short-range mobility layer is generated synthetically by relying on the “gravity law”  or the more recent “radiation law”  both calibrated using real data \cite{Simini2012}. 

The model is calibrated to realistically describe the evolution of the COVID-19 pandemic as detailed in \cite{chinazzi2020effect,davis2020estimating}.
Lastly, GLEAM is stochastic and produces an ensemble of possible epidemic outcomes for each set of initial conditions. To account for the potentially different reporting levels of the states, a free parameter Infection Fatality Rate (IFR) multiplier is added to each model. To calibrate and select the most reasonable outcomes, we filter the models by the latest hospitalization trends and confirmed cases trends, and then we select and weight the filtered models using Akaike Information Criterion \cite{zhang2017forecasting}. The forecast of the evolution of the epidemic is formed by the final ensemble of the selected models.

\paragraph{DCRNN setup}
The model has one hidden layer of RNN with 8 units to overcome the overfitting problem. The filter type is Laplacian and the diffusion step is 1. The base learning rate of DCRNN is $1e^{-2}$ and decay to $1e^{-3}$ at epoch 13 with Adam optimizer \cite{kingma2014adam}. We have a strict early stopping policy to deal with the overfitting problem. The training stops as the validation error does not improve for three epochs after epoch 13.

\paragraph{Bootstrap}
For Bootstrap method, due to the small sample we have (typically 25), we only randomly dropped 1 training data while keeping the original validation and testing data. We obtain 25 samples for constructing mean prediction and confidence interval. 
\paragraph{Quantile regression}
We apply pinball loss function \cite{Koenker1978,Koenker2005} to train three different quantiles for the corresponding confidence interval (CI), e.g., (0.025, 0.5, 0.975) for 95\%. The model and learning rate setup is the same as DCRNN. The result for comparison averages the performance of 10 trails. 

\paragraph{SQ regression}
We use spline quantile function to approximate a quantile function and use the CRPS as the loss function. For every point prediction, there are 11 trained parameters to construct the quantile function. The 1\ts{st} parameter is the intercept term. The next 5 can be transformed to the slopes of 5 line segments. The last 5 can be transformed to a vector of the 5 knots' positions. The model and learning rate setup is the same as DCRNN. The result for comparison averages the performance of 10 trails. 

\paragraph{MIS regression}
We combine MAE with MIS and directly minimize this loss function. The model and learning rate setup is the same as DCRNN. The result for comparison averages the performance of 10 trails. 
\paragraph{MC Dropout}
The training process of MC Dropout is the same as DCRNN. We implement the algorithm provided by \cite{zhu2017deep} and simplify the model by only considering the model uncertainty. We apply random dropout through the testing process with 5\% drop rate and iterate 300 times to achieve a stable prediction. The result for comparison averages the performance of 10 trails. 

\paragraph{SG-MCMC}
To generate samples of model parameters $\theta$ (with a slight abuse of notation) according to SG-MCMC. In this task of complex time-series, the temperature of system is hard to be tuned. Stochastic Gradient Nose-Hoover Thermostats (SGNHT) could avoid this problem in training. We first denote the loss function (or the negative log-likelihood) over a minibatch of data as $\widetilde{\mathcal{L}}(\theta)$.
We then introduce hyper-parameters including the diffusion coefficients $A$ and the learning rate $h$ and make use of auxiliary variables $p\in\mathbb{R}^{d}$ and $\xi\in\mathbb{R}$ in the algorithm.
We randomly initialize $\theta$, $p$, and $\xi$ and update according to the following update rule.
\begin{align}
\label{eq:SGNHT}
\left\{
\begin{array}{l}
    \theta_{k+1} = \theta_k + p_k h  \\
    p_{k+1} = p_k - \nabla \widetilde{\mathcal{L}}(\theta) h - \xi_k p_k h + \mathcal{N}(0,2A h) \\
    \xi_{k+1} = \xi_k + \lrp{ \frac{p_k^\rT p_k}{d} - 1 } h.
\end{array}
\right.
\end{align}
Upon convergence of the above algorithm at $K$-th step, $\theta_K$ follows the distribution of the posterior.
We run parallel chains to generate different samples according to the posterior and quantify the predictions uncertainty.\\
The learning rate of SGNHT is $5e^{-4}$, and we selected a Gaussian prior $\normal(0, 2.0)$ with random initialization as $\NORMAL(0, 0.05)$. We apply training for 800 epochs, and it early stops as long as the validation error does not improve for 50 epochs. Our result is averaged from 25 posterior samples. 

\section{Performance of Ensemble DeepGLEAM model}
In addition to autoregressive forecasting, we also build an ensemble DeepGLEAM model by training based on the ensembled data along the forecasting horizon as the output.
We use the ensemble DeepGLEAM model which trains four different models separately and predicts the future four weeks death prediction respectively. Table \ref{tb:ensemble_table} shows the performance of the UQ methods. We note here the autoregressive DeepGLEAM outperforms Ensemble DeepGLEAM in both Frequentist and Bayesian UQ methods in the first three weeks. For longer time-series predictions, the accumulated bias from autoregressive model may negatively affect the performance compared to the separately trained ensemble model. In the context of COVID-19 prediction, autoregressive DeepGLEAM is a better choice. 

\begin{table}[h]
\centering
\label{tab:covid_comparison}
\resizebox{\columnwidth}{!}{
\begin{tabular}{c|c|cccccc}
\toprule
$T$ & Metric & Bootstrap & Quantile & SQ & MAE-MIS & MC Drpoout & SG-MCMC \\ 
\midrule
\multirow{2}{*}{1W}  & RMSE & 70.73 & 70.26 & 73.82 & 76.32 & 68.58 & \textbf{62.42}    \\ 
& MIS (95$\%$ CI) & 1216.63 & 430.72 & 1299.78 & \textbf{424.49} & 831.86 & 715.40  \\ 
\cline{1-8}
\multirow{2}{*}{2W} & RMSE & 61.69  & 60.94 & 65.44 & 65.39 & 59.27 & \textbf{51.21}    \\ 
& MIS (95$\%$ CI) & 1135.01 & \textbf{331.94} & 1272.82 & 368.20 & 796.54 & 727.27   \\ 
\cline{1-8}
\multirow{2}{*}{3W} & RMSE & 73.27 & 71.72 & 73.88 & 74.95 & 70.32 & \textbf{66.72}  \\ 
& MIS (95$\%$ CI) & 1464.52 & \textbf{392.66} & 1559.62 & 416.23 & 997.65 & 930.55  \\ 
\cline{1-8}
\multirow{2}{*}{4W} & RMSE & 72.41 & 73.44 & 70.44 & 75.16 & 72.24 & \textbf{68.40}  \\ 
& MIS (95$\%$ CI) & 1482.08 & \textbf{418.77} & 1586.95 & 468.49 & 1034.68 & 971.55 \\ 
\bottomrule
\end{tabular}}
\caption{Performance comparison of different approaches on Ensemble DeepGLEAM Model for COVID-19 mortality forecasting. }
\label{tb:ensemble_table}
\end{table}

\end{document}